\documentclass[journal]{IEEEtran}

\usepackage[utf8]{inputenc}
\usepackage[T1]{fontenc}
\usepackage{lmodern} 
\usepackage{amsmath, amssymb} 
\usepackage{graphicx} 
\usepackage{hyperref} 
\usepackage{cite} 
\usepackage{times}
\usepackage{adjustbox} 

\title{CNN-LSTM Hybrid Architecture for Over-the-Air Automatic Modulation Classification Using SDR
}

\author{Dinanath Padhya, Krishna Acharya, Bipul Kumar Dahal, Dinesh Baniya Kshatri%
\thanks{Dinanath Padhya, Krishna Acharya, and Bipul Kumar Dahal are with the Department of Electronics and Computer Engineering, Thapathali Campus, Institute of Engineering, Tribhuvan University Emails: \{dinanath.078bei010, krishna.078bei020, bipul.078bei009\}@tcioe.edu.np}%
\thanks{Dinesh Baniya Kshatri is the supervisor for this work.}
}

\begin{document}

\maketitle

\begin{abstract}
    Automatic Modulation Classification (AMC) is a core technology for future wireless communication systems, enabling the identification of modulation schemes without prior knowledge. This capability is essential for applications in cognitive radio, spectrum monitoring, and intelligent communication networks. We propose an AMC system based on a hybrid Convolutional Neural Network (CNN) and Long Short-Term Memory (LSTM) architecture, integrated with a Software Defined Radio (SDR) platform. The proposed architecture leverages CNNs for spatial feature extraction and LSTMs for capturing temporal dependencies, enabling efficient handling of complex, time-varying communication signals. The system's practical ability was demonstrated by identifying over-the-air (OTA) signals from a custom-built FM transmitter alongside other modulation schemes. The system was trained on a hybrid dataset combining the RadioML2018 dataset with a custom-generated dataset, featuring samples at Signal-to-Noise Ratios (SNRs) from 0 to 30~dB. System performance was evaluated using accuracy, precision, recall, F1 score, and the Area Under the Receiver Operating Characteristic Curve (AUC-ROC). The optimized model achieved 93.48\% accuracy, 93.53\% precision, 93.48\% recall, and an F1 score of 93.45\%. The AUC-ROC analysis confirmed the model's discriminative power, even in noisy conditions. This paper's experimental results validate the effectiveness of the hybrid CNN-LSTM architecture for AMC, suggesting its potential application in adaptive spectrum management and advanced cognitive radio systems.
\end{abstract}


\section{Introduction}
Automatic Modulation Classification (AMC) is a vital technology in modern wireless communication systems, enabling modulation identification without prior signal knowledge. This capability facilitates efficient spectrum use, interference detection, and enhanced security in applications like cognitive radio networks and defense systems. The rapid expansion of connected devices, particularly within the Internet of Things (IoT) and smart cities, has significantly increased radio frequency spectrum congestion. Traditional AMC methods, such as likelihood-based approaches, are often incapable of handling real-world signal variations like noise, fading, and Doppler shifts. To address these limitations, recent deep learning research has produced hybrid models combining Convolutional Neural Networks (CNNs) and Long Short-Term Memory (LSTM) networks. These models can extract both spatial and temporal signal features, resulting in improved classification accuracy and robustness under challenging channel conditions.

\section{Related Work}
Early Automatic Modulation Classification (AMC) techniques mainly used maximum likelihood (ML) and feature-based (FB) methods. While ML methods are optimal theoretically, their high computational complexity limits real-time use \cite{Grimaldi2007An,Ramkumar2009Automatic}. FB methods rely on handcrafted features, requiring expert knowledge and often struggling in complex environments \cite{Grimaldi2007An,Meng2018Automatic}. Cyclostationary-based and approximate entropy features improved performance in low SNR conditions by exploiting signal periodicities \cite{dobre2015,Ramkumar2009Automatic}.

The introduction of deep learning (DL), especially Convolutional Neural Networks (CNNs), marked a major shift by automatically extracting features and significantly improving classification accuracy and speed \cite{Meng2018Automatic}. Hybrid models combining CNNs with Long Short-Term Memory (LSTM) networks effectively capture both spatial and temporal dependencies, enhancing performance in dynamic environments \cite{lstm,zhang2020cnn}. Complex-valued networks further improved handling of inherently complex radio signals \cite{Tu2020Complex-Valued}.

Recent work focuses on addressing intra-class diversity and noise robustness using multi-scale networks and denoising autoencoders \cite{Zhang2021A,Wu2017Robust,An2023Robust}. Advanced architectures combining CNNs, LSTMs, and transformers capture both spatial and sequential signal features for better adaptability to channel impairments \cite{Huynh-The2021Automatic}. Manifold learning and multimodal fusion approaches, integrating time, frequency, and phase information, have demonstrated superior classification accuracy in challenging scenarios \cite{Li2012Manifold,Qi2021Automatic,Zhang2019Automatic,Deng2023Cross-Domain}.

Furthermore, unsupervised and semi-supervised learning techniques, including generative adversarial networks (GANs), have been explored to mitigate the need for large labeled datasets. Transfer learning and domain adaptation methods improve generalization across varying channel conditions \cite{Deng2023Cross-Domain}. Effective signal representation and preprocessing methods, such as time-frequency analysis, cooperative contrast learning, and multi-feature fusion, play critical roles in enhancing classification performance \cite{Liu2022Wireless,Bai2024Achieving,Huang2019Automatic,Zheng2022Fine-Grained}.

Finally, real-time AMC implementations leveraging FPGA, RFSoC hardware, and SDR integrated with DL architectures have enabled low-latency, high-performance modulation classification systems suited for practical deployment \cite{realtimeamc2029trigdell,multiskip2020lin}.

Despite these advancements, existing literature predominantly relies on synthetic datasets and simulation-based analysis, often overlooking the complex channel impairments and signal distortions inherent in real-world environments. Furthermore, architectures utilizing CNNs or LSTMs separately often struggle to capture both spatial and temporal dependencies simultaneously, limiting robust classification under varying SNR conditions. We address these gaps by proposing a novel hybrid CNN-LSTM architecture that efficiently integrates both feature types to outperform standalone baselines. Uniquely, we move beyond simulation by implementing a custom SDR testbed to evaluate the model's performance on over-the-air (OTA) signals, ensuring practical, real-world applicability.

\section{Methodology}
\begin{figure}[h]
    \centering
    \includegraphics[width=0.48\textwidth]{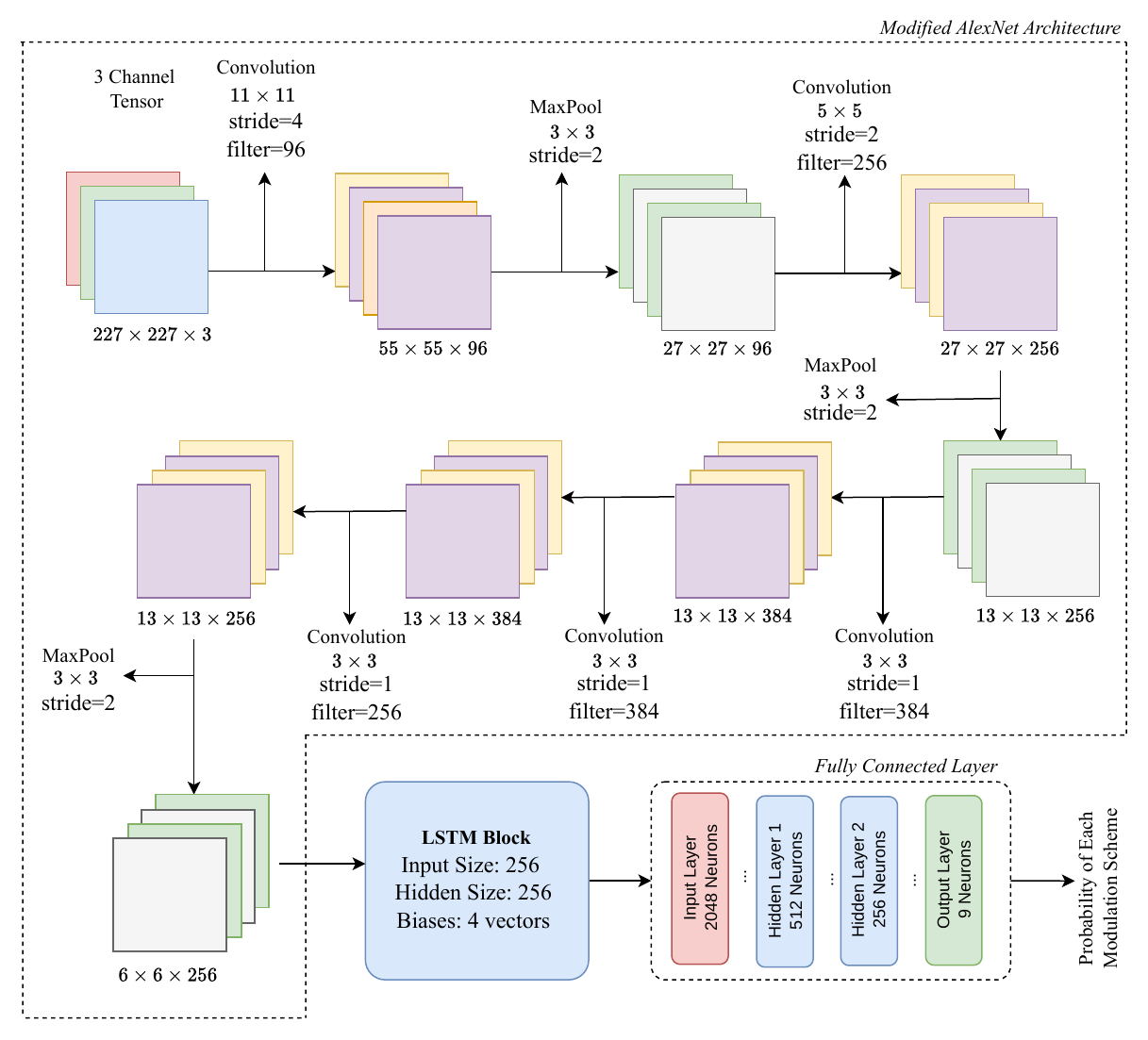}
    \caption{CNN-LSTM hybrid architecture: AlexNet-based feature extractor, LSTM for temporal modeling, and classification head}
    \label{fig:model_architecture}
\end{figure}

As depicted in Figures~\ref{fig:input_preprocessing} and~\ref{fig:model_architecture}, the AMC system consists of six major components: signal generation and OTA transmission using GNU Radio and SDR hardware, I/Q signal reception using RTL-SDR, transformation of the signal into amplitude, phase, and I/Q domains, CNN-based spatial feature extraction, temporal modeling through LSTM, and final classification into modulation categories.

\subsection{Signal Acquisition Using SDR}
The SDR-based signal acquisition begins with the generation of modulated signals in GNU Radio. Multiple modulation schemes such as BPSK, QPSK, AM, FM, 8PSK, 16QAM, and GMSK are synthesized in real time. These signals are either looped back virtually or transmitted over-the-air via an SDR transmitter. The receiving end utilizes RTL-SDR hardware to capture the signals. This device comprises an RF front-end for downconversion, an ADC for digitization, and a USB interface that streams raw I/Q samples to the host PC for downstream processing.

\subsection{Dataset Generation}
The AMC model is trained on a hybrid dataset comprising both publicly available and custom-generated signals. The publicly available portion comes from the RadioML2018.01a dataset, which contains labeled I/Q samples across multiple modulation schemes under varying channel conditions. To complement this, we constructed an additional dataset using GNU Radio, where signals were synthetically generated for a similar set of modulation types. This custom dataset includes both clean signals and noisy variants injected with Additive White Gaussian Noise (AWGN).

Signal modulation was carried out using GNU Radio flowgraphs, supporting schemes such as m-PSK, AM, FM, m-QAM, GMSK, and GFSK. The modulated signals were either virtually looped back or transmitted over-the-air using Software Defined Radios (SDRs). Reception was handled by an RTL-SDR device, which performs RF downconversion and digitization, streaming the raw I/Q samples to a host machine.

\begin{figure}[!h]
    \centering
    \includegraphics[width=0.45\textwidth]{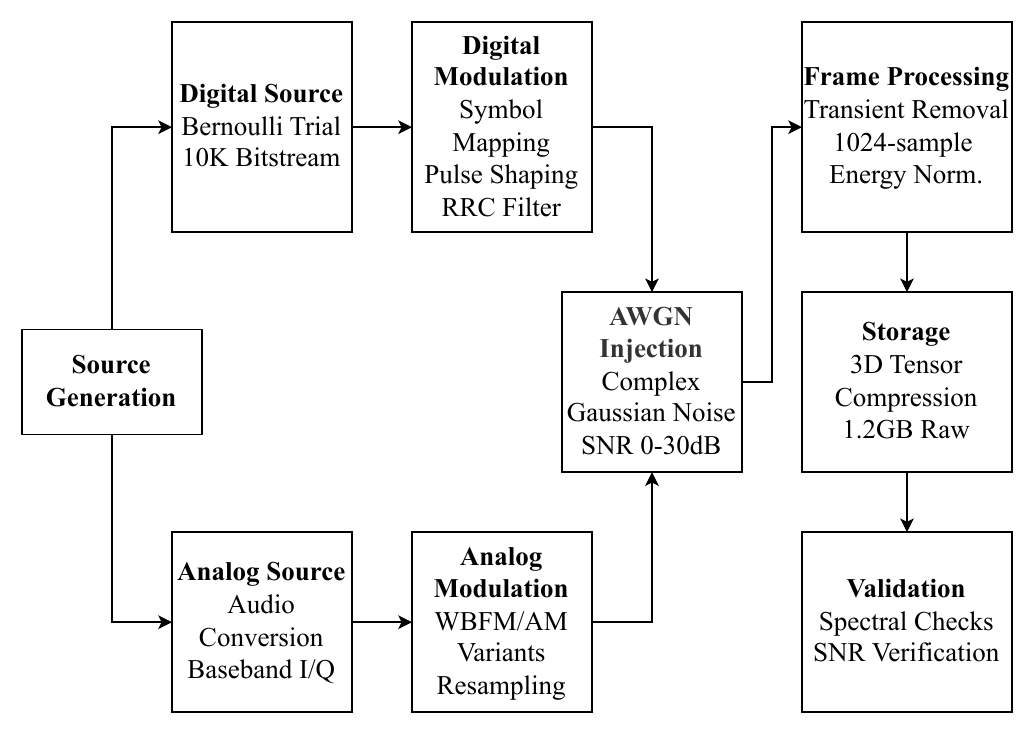}
    \caption{Dataset generation and validation pipeline implemented in GNU Radio for creating both clean and AWGN-corrupted signal samples across multiple modulation schemes}
    \label{fig:dataset_generation_valid_pipeline}
\end{figure}

The complete dataset generation and validation process is illustrated in Figure~\ref{fig:dataset_generation_valid_pipeline}, which shows both the signal synthesis flow and the validation pipeline used to ensure signal integrity and class balance across modulation types.

\subsection{Preprocessing and Input Representation}
After acquisition, each I/Q sample stream is processed into a format suitable for deep learning. Every signal instance, composed of 1024 I/Q samples, is segmented into eight overlapping windows of 224 samples. The 50\% overlap ensures better temporal continuity and increases the effective training sample size.

For each segmented window, three signal representations are derived. First, the amplitude is computed as:
\begin{equation}
    A[n] = \sqrt{I[n]^2 + Q[n]^2}
\end{equation}
which captures the signal envelope. Second, the phase is extracted as:
\begin{equation}
    \phi[n] = \arctan\left(\frac{Q[n]}{I[n]}\right)
\end{equation}
which captures frequency variations. Third, the raw I/Q values are retained to preserve the baseband vector information.

\begin{figure}[h]
    \centering
    \includegraphics[width=0.45\textwidth]{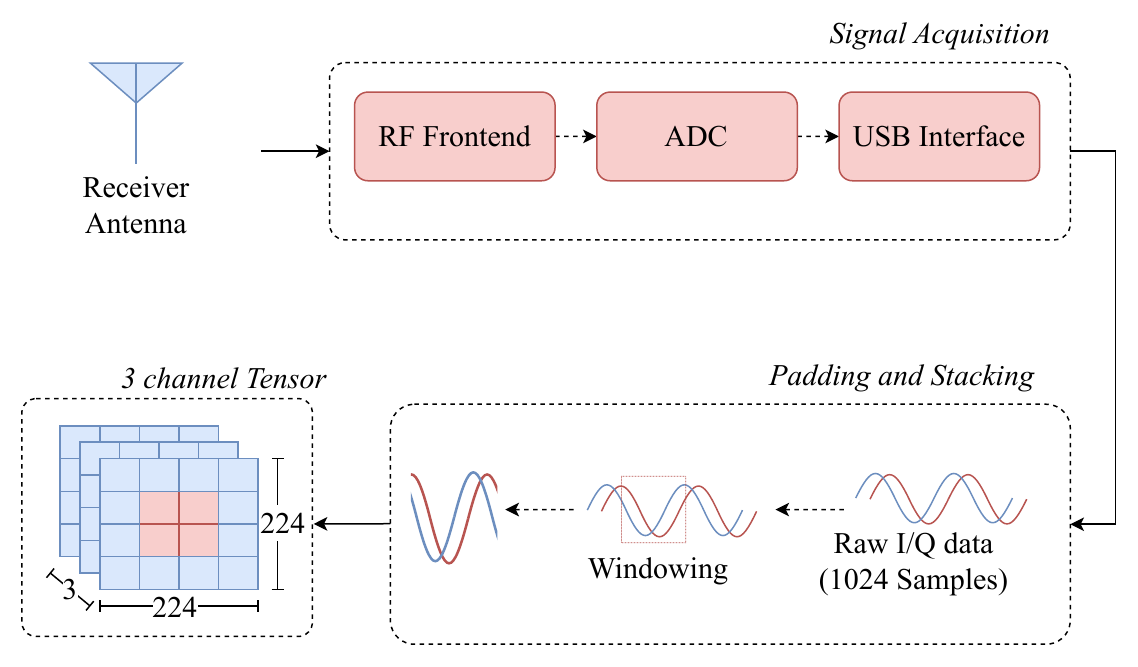}
    \caption{Input preprocessing pipeline showing transformation of I/Q samples into amplitude, phase, and I/Q representations for CNN-LSTM model input}
    \label{fig:input_preprocessing}
\end{figure}

\begin{figure}[!htbp]
    \centering
    \includegraphics[width=0.8\linewidth]{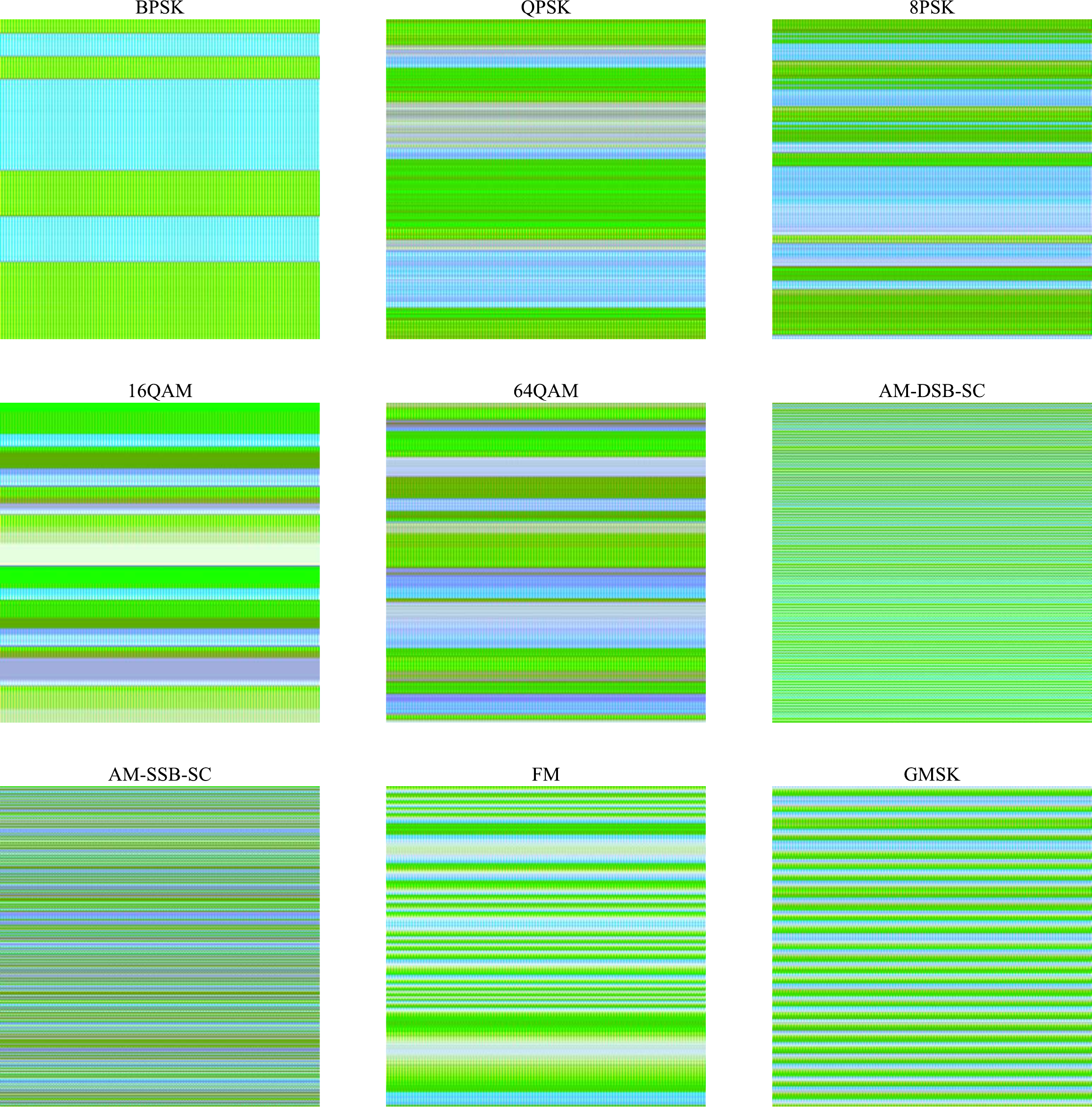}
    \caption{Preprocessed RGB image of Amplitude, I/Q and Phase}
    \label{fig:final_representation}
\end{figure}

Each of these one-dimensional sequences is reshaped into a \(224 \times 224\) two-dimensional matrix using repetition and tiling operations. These matrices of amplitude, phase, and I/Q are then stacked along the third axis to form a unified \(224 \times 224 \times 3\) tensor. This tensor resembles an RGB image where each channel encodes a distinct physical property of the signal. This 3-channel representation is used as input to the CNN backbone of the model. The complete preprocessing pipeline from I/Q samples to feature tensor construction is visualized in Figure~\ref{fig:input_preprocessing}, while an example of the final tensor is shown in Figure~\ref{fig:final_representation}.

\subsection{Modified AlexNet}
The classification model integrates a spatial feature extractor based on a modified AlexNet and a temporal model based on an LSTM network. This hybrid structure enables joint exploitation of spatial patterns across each window and sequential dependencies between windows. The convolutional component of the network is based on a modified version of the AlexNet architecture, adapted to process \(224 \times 224 \times 3\) tensors representing amplitude, phase, and I/Q information. To reduce parameter overhead and promote generalization, the original fully connected layers from AlexNet are removed. The CNN comprises five convolutional layers with ReLU activations. The first and second layers are followed by Local Response Normalization (LRN) and max-pooling to stabilize learning and reduce spatial dimensions.

Specifically, Conv1 uses an \(11 \times 11\) kernel with stride 4 and outputs 96 feature maps. Conv2 applies a \(5 \times 5\) kernel with 256 filters. Conv3 to Conv5 employ \(3 \times 3\) kernels with 384, 384, and 256 filters, respectively. The final output of the convolutional stack is a feature map of size \(13 \times 13 \times 256\).

To compress this output into a compact representation, global average pooling (GAP) is applied, producing a 256-dimensional feature vector.

\subsection{Temporal Modeling via LSTM}
The LSTM module processes sequences of 256-dimensional CNN feature vectors extracted from eight overlapping windows of each I/Q signal. The input to the LSTM has shape \((B, 8, 256)\), where \(B\) denotes the batch size. The model uses a single unidirectional LSTM layer with a hidden size of 256 and includes a dropout of 0.6 to prevent overfitting.

The LSTM captures temporal dynamics across successive windows by maintaining a memory cell and using gates to regulate information flow. The internal LSTM operations for each timestep \(t\) are governed by:
\begin{align}
    f_t         & = \sigma(W_f [h_{t-1}, x_t] + b_f)          \\
    i_t         & = \sigma(W_i [h_{t-1}, x_t] + b_i)          \\
    \tilde{c}_t & = \tanh(W_c [h_{t-1}, x_t] + b_c)           \\
    c_t         & = f_t \odot c_{t-1} + i_t \odot \tilde{c}_t \\
    o_t         & = \sigma(W_o [h_{t-1}, x_t] + b_o)          \\
    h_t         & = o_t \odot \tanh(c_t)
\end{align}
Here, \(f_t, i_t, o_t\) represent the forget, input, and output gates; \(c_t\) is the cell state, and \(h_t\) is the hidden state at timestep \(t\).

The final LSTM output is flattened into a 2048-dimensional vector to be fed into the classification network.

\subsection{Fully Connected Classification}
The flattened output from the LSTM is passed through a fully connected classifier to predict the modulation type. The classification head consists of two dense layers with 512 and 256 neurons, respectively. Both layers use leaky ReLU activations to maintain gradient flow in the presence of negative activations. The final output layer is a softmax layer with 9 units, corresponding to the number of modulation classes in the dataset.

The training process optimizes a categorical cross-entropy loss, defined as:
\begin{equation}
    \mathcal{L}_{\text{CE}} = -\sum_{i=1}^{N} \sum_{j=1}^{C} y_{i,j} \log(\hat{y}_{i,j})
\end{equation}
where \(N\) is the number of samples, \(C\) is the number of classes, \(y_{i,j}\) is the true label, and \(\hat{y}_{i,j}\) is the predicted probability.

We ran several trainings to tune hyperparameters (dropout rate, number of neurons in the first hidden layer, learning rate) in a systematic way in order to get the best generalization on the validation set. The Adam optimizer was used. Early stopping on validation loss was performed to avoid unnecessary training and to lessen the danger of overfitting. The data was split into 80\% for training, 10\% for validation, and 10\% for testing. The final model evaluation was done by metrics like accuracy, precision, recall, F1-score, and AUC-ROC together with confusion matrix visualization to measure the performance of class-wise prediction.

\begin{table}[ht]
    \centering
    \caption{Hyperparameter settings tested during model optimization}
    \label{tab:hyperparams_initial}
    \footnotesize
    \setlength{\tabcolsep}{4pt}
    \begin{tabular}{|c|c|c|c|}
        \hline
        \textbf{Trial} & \textbf{Drop Factor} & \textbf{L1} & \textbf{LR ($\times 10^{-4}$)} \\ \hline
        1-3            & 0.4                  & 1024        & 1.0, 0.5, 1.5                  \\ \hline
        4-6            & 0.4                  & 512         & 1.0, 0.5, 1.5                  \\ \hline
        7-9            & 0.5                  & 1024        & 1.0, 0.5, 1.5                  \\ \hline
        10-12          & 0.5                  & 512         & 1.0, 0.5, 1.5                  \\ \hline
        13-15          & 0.6                  & 1024        & 1.0, 0.5, 1.5                  \\ \hline
        16-18          & 0.6                  & 512         & 1.0, 0.5, 1.5                  \\ \hline
    \end{tabular}
\end{table}

\section{Results and Discussion}
We conducted various trials to evaluate the performance of the proposed system, focusing on different batch sizes and hyperparameter tuning as described in Table~\ref{tab:hyperparams_initial}.

\subsection{Model Training and Validation Performance}
First, we trained the model with batch sizes of 16 and 32 and then performed hyperparameter tuning to further improve the model performance.

\begin{figure}[!t]
    \centering
    \includegraphics[width=\columnwidth]{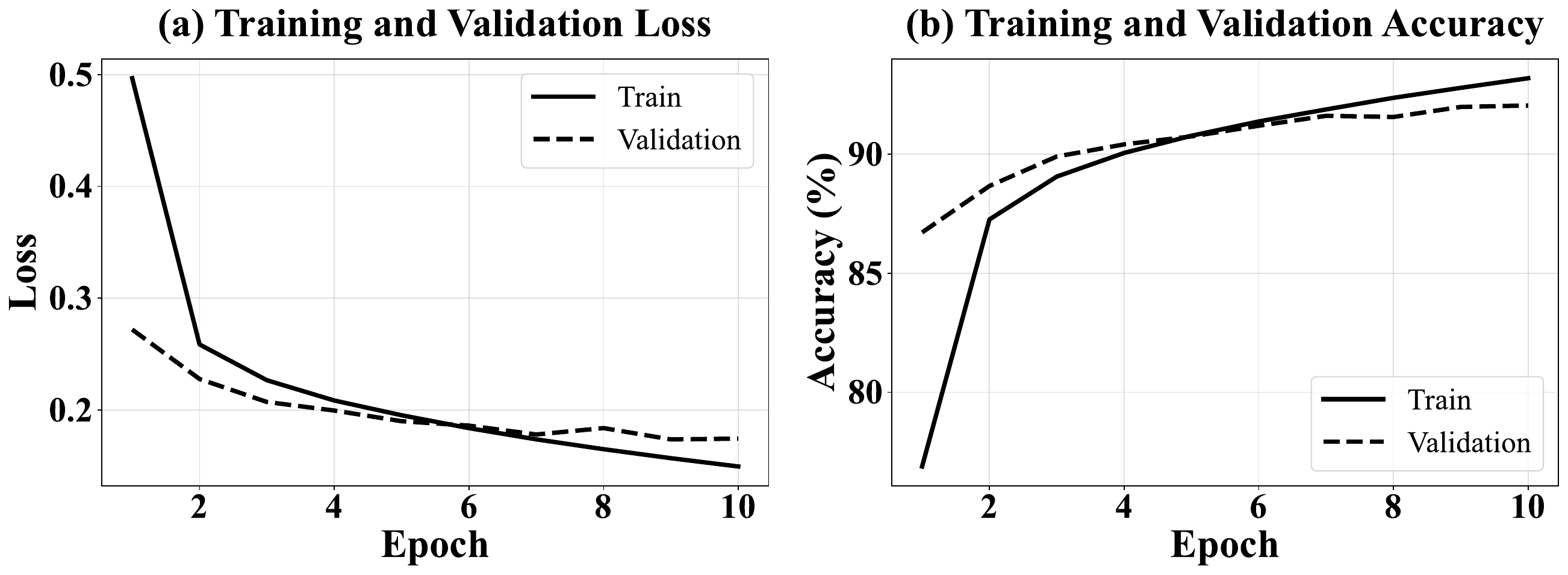}
    \caption{Training and validation curves over 10 epochs for the optimized CNN-LSTM model}
    \label{fig:train_val_loss_tuned}
\end{figure}

Based on the hyperparameter tuning results, we selected the best model with a validation loss of 0.251 and a training loss of 0.251 trained for 5 iterations over 7352 seconds. The best parameters were a drop factor of 0.6, number of neurons in first hidden layer of 512, and a learning rate of 0.00015, which we used for the final model training. We trained the model with a batch size of 32, achieving a final accuracy of 93.48\% on the test set. The training and validation curves for the optimized model are shown in Figure~\ref{fig:train_val_loss_tuned}.

\begin{figure}[!t]
    \centering
    \includegraphics[width=\columnwidth]{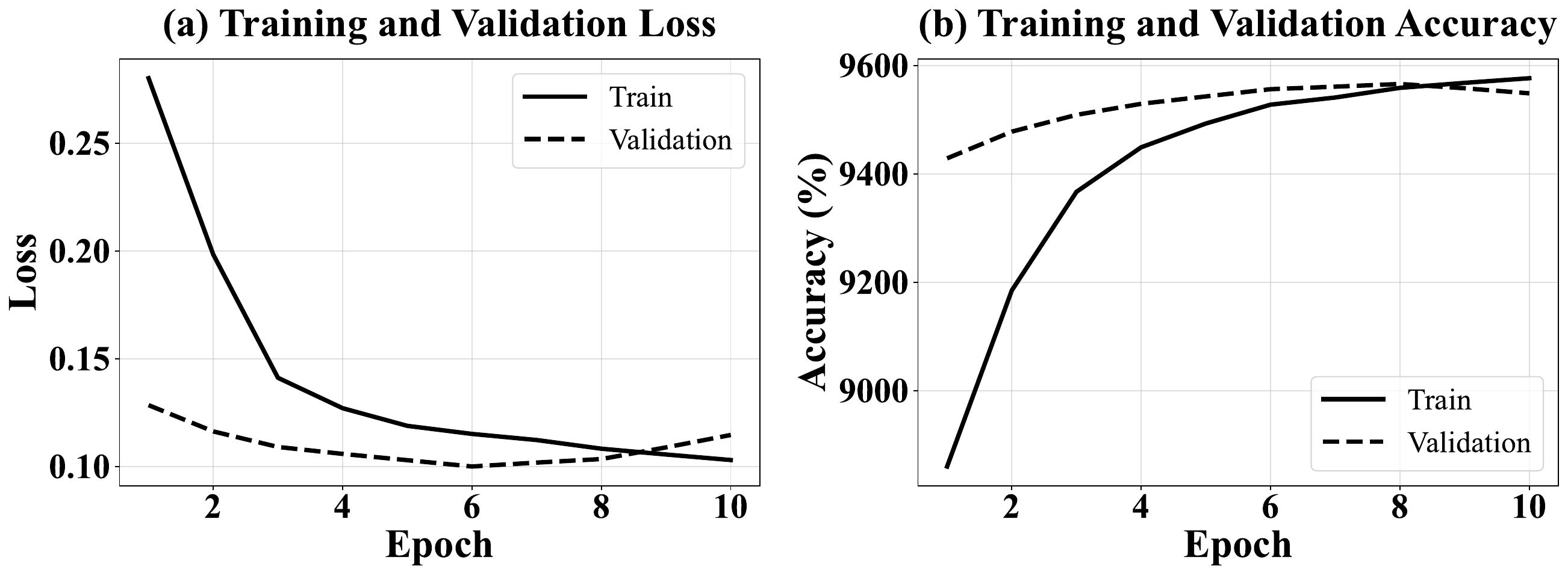}
    \caption{Training and validation curves over 10 epochs for attention based model}
    \label{fig:train_val_loss_attention}
\end{figure}

To further optimize the model, we conducted a comparative experiment by replacing the LSTM's hidden state single flattening mechanism with a single-headed temporal attention layer. This alternative approach was designed to identify and focus on the most salient time steps within the 8-sequence windows. Instead of flattening the 8 hidden state outputs (an \(8 \times 256\) matrix) into a 2048-dimensional vector, the attention model computed a weighted average of these outputs. This process compressed the sequence into a single 256-dimensional vector before classification. This modification resulted in degraded performance, with accuracy dropping to 93.34\% and the F1-score decreasing to 93.22\%, as illustrated in Figure~\ref{fig:train_val_loss_attention}.

\subsection{Confusion Matrix Analysis}
\begin{figure}[!t]
    \centering
    \includegraphics[width=\columnwidth]{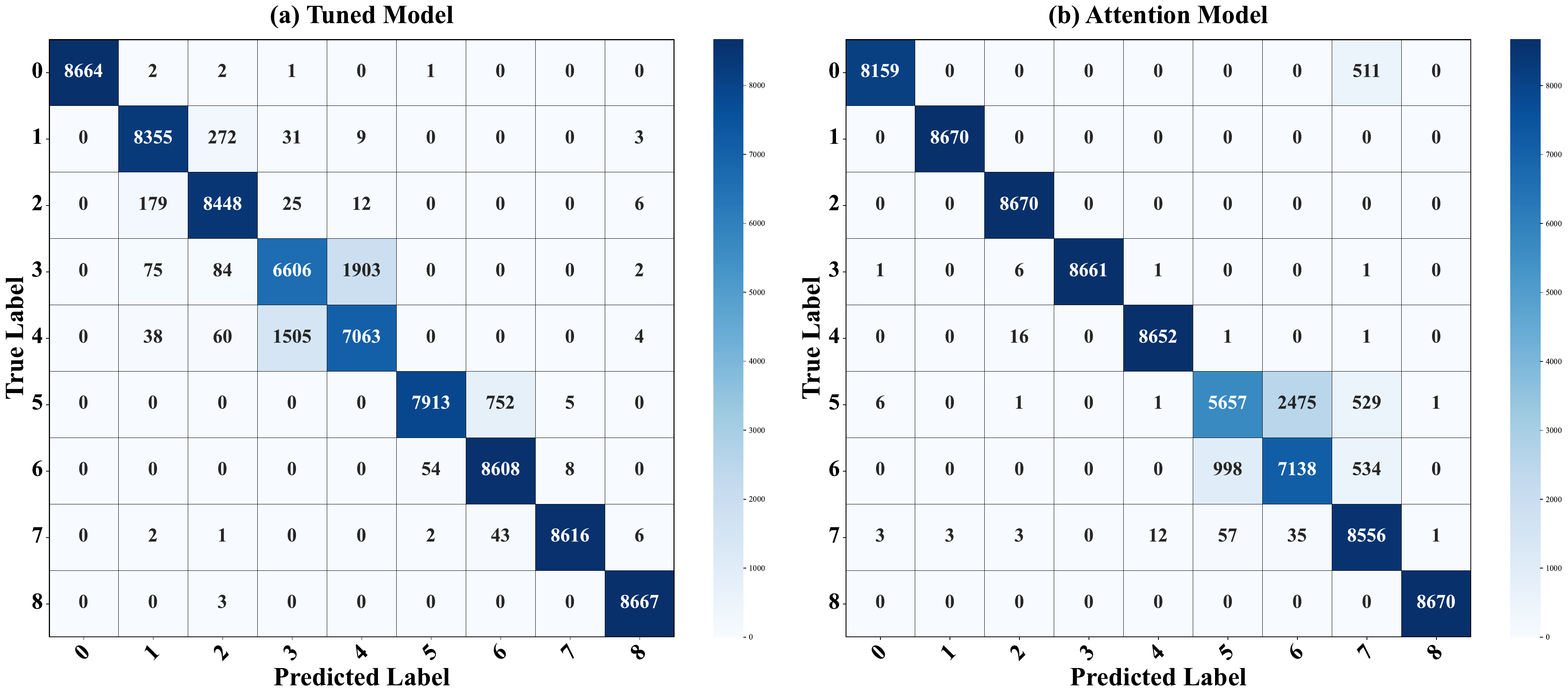}
    \caption{Confusion matrices comparing the classification performance of the Proposed CNN-LSTM (a) and the Attention model (b) across nine modulation schemes}
    \label{fig:confusion_matrix_comparison}
\end{figure}

The confusion matrices presented in Figure~\ref{fig:confusion_matrix_comparison} provide a comparative analysis of the model's classification capabilities. Figure~\ref{fig:confusion_matrix_comparison}(a) illustrates the proposed optimized model, while Figure~\ref{fig:confusion_matrix_comparison}(b) represents the comparative attention-based model.

Across the proposed model in Figure~\ref{fig:confusion_matrix_comparison}(a), there is strong diagonal dominance, indicating high overall accuracy. The primary misclassifications remain between 16QAM and 64QAM (Classes 3 and 4), which is expected due to the inherent similarity of their constellation structures under noisy conditions.

In contrast, the attention-based model in Figure~\ref{fig:confusion_matrix_comparison}(b) reveals a significant performance degradation, particularly in distinguishing analog modulation schemes. Substantial confusion is evident between AM-DSB-SC and AM-SSB-SC (Classes 5 and 6), indicated by the high off-diagonal values in those regions. This suggests that the attention mechanism struggles to preserve the fine-grained spectral features required to separate these closely related analog signals, a limitation that the proposed optimized model successfully overcomes.

We evaluated the model's accuracy, precision, recall, and F1 score across different batch sizes and hyperparameter configurations. We summarize the results in Table~\ref{tab:model_performance}, showing that the model maintains strong classification performance across configurations, with notable improvement after tuning. The optimized model achieved an accuracy of 93.48\%, precision of 93.53\%, recall of 93.48\%, and F1 score of 93.45\%, significantly outperforming both the baseline and the single-headed attention configurations.

\begin{table}[!t]
    \centering
    \caption{Performance Metrics Comparison Across Different Model Configurations and Architectures}
    \label{tab:model_performance}
    \begin{tabular}{|c|c|c|c|c|}
        \hline
        \textbf{Configuration}  & \textbf{Accuracy} & \textbf{Precision} & \textbf{Recall} & \textbf{F1 Score} \\
                                & \textbf{(\%)}     & \textbf{(\%)}      & \textbf{(\%)}   & \textbf{(\%)}     \\
        \hline
        Batch Size 16           & 91.46             & 91.46              & 91.46           & 91.33             \\
        \hline
        Batch Size 32           & 91.34             & 91.47              & 91.34           & 91.30             \\
        \hline
        Batch Size 32 (Tuned)   & 93.48             & 93.53              & 93.48           & 93.45             \\
        \hline
        Single Headed Attention & 93.34             & 93.56              & 93.34           & 93.22             \\
        \hline
    \end{tabular}
\end{table}

\subsection{Receiver Operating Characteristic (ROC) Analysis}
\begin{figure}[!t]
    \centering
    \includegraphics[width=\columnwidth]{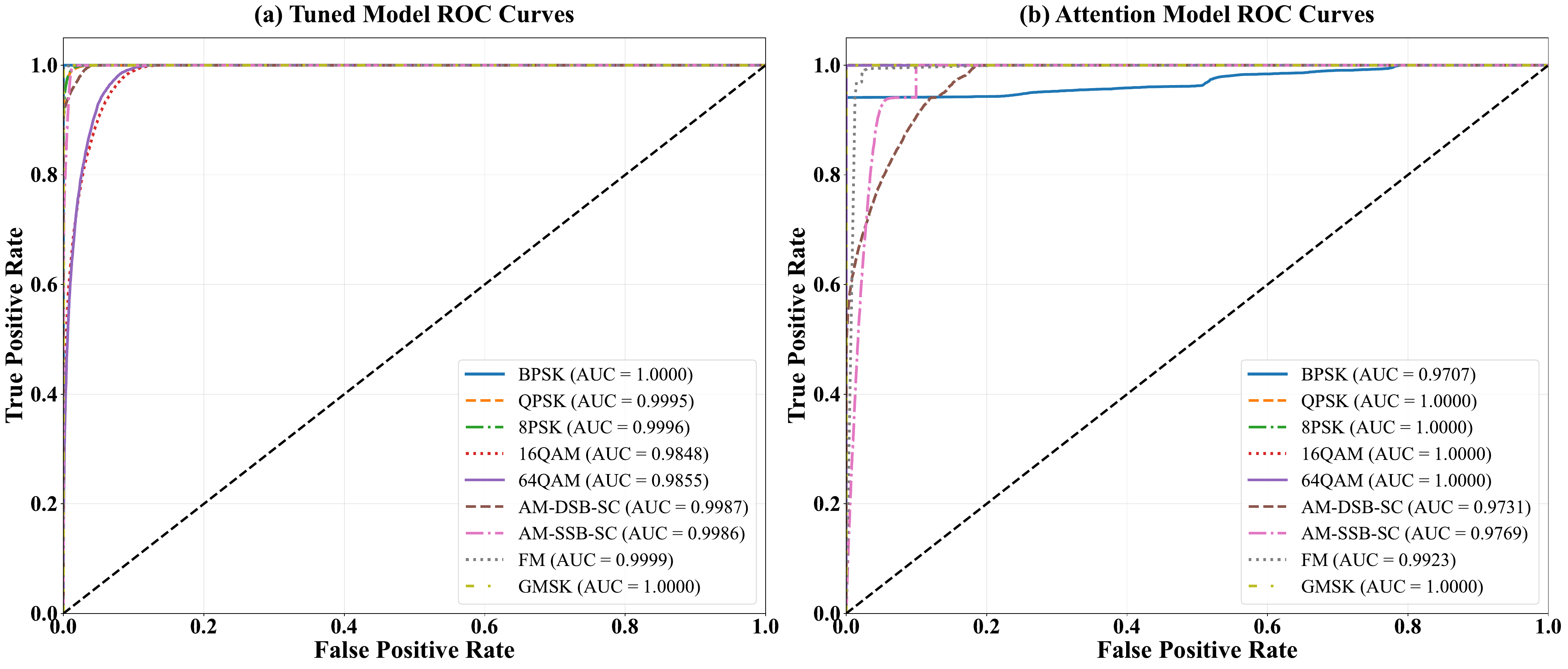}
    \caption{Comparative ROC curves showing the superior discriminative capability of the Proposed CNN-LSTM model (a) over the Attention model (b)}
    \label{fig:roc_curve_comparison}
\end{figure}

The ROC curves in Figure~\ref{fig:roc_curve_comparison} demonstrate the model's outstanding ability in identifying modulation classes. As depicted in Figure~\ref{fig:roc_curve_comparison}(a), the Tuned Model attains Area Under the Curve (AUC) almost 1.00 for most classes, thus the classification is reliable. That means that BPSK, QPSK, 8PSK, FM, and GMSK can be perfectly separated. Although 16QAM and 64QAM have slightly lowered AUC values (0.9848 and 0.9855) because of their tightly packed constellation structures, the model is still highly discriminative.

Conversely, the Attention Model (Figure~\ref{fig:roc_curve_comparison}(b)) shows a clear performance decline. Specifically, the ROC curves for the two analog modulations (AM-DSB-SC and AM-SSB-SC) deviate significantly from the top-left corner, with their AUC values falling to 0.97. This demonstrates that the proposed flattening design preserves essential temporal features for analog signals, which are otherwise lost during the attention-based compression.

\subsection{Precision-Recall and F1 Stability}
\begin{figure}[!t]
    \centering
    \includegraphics[width=\columnwidth]{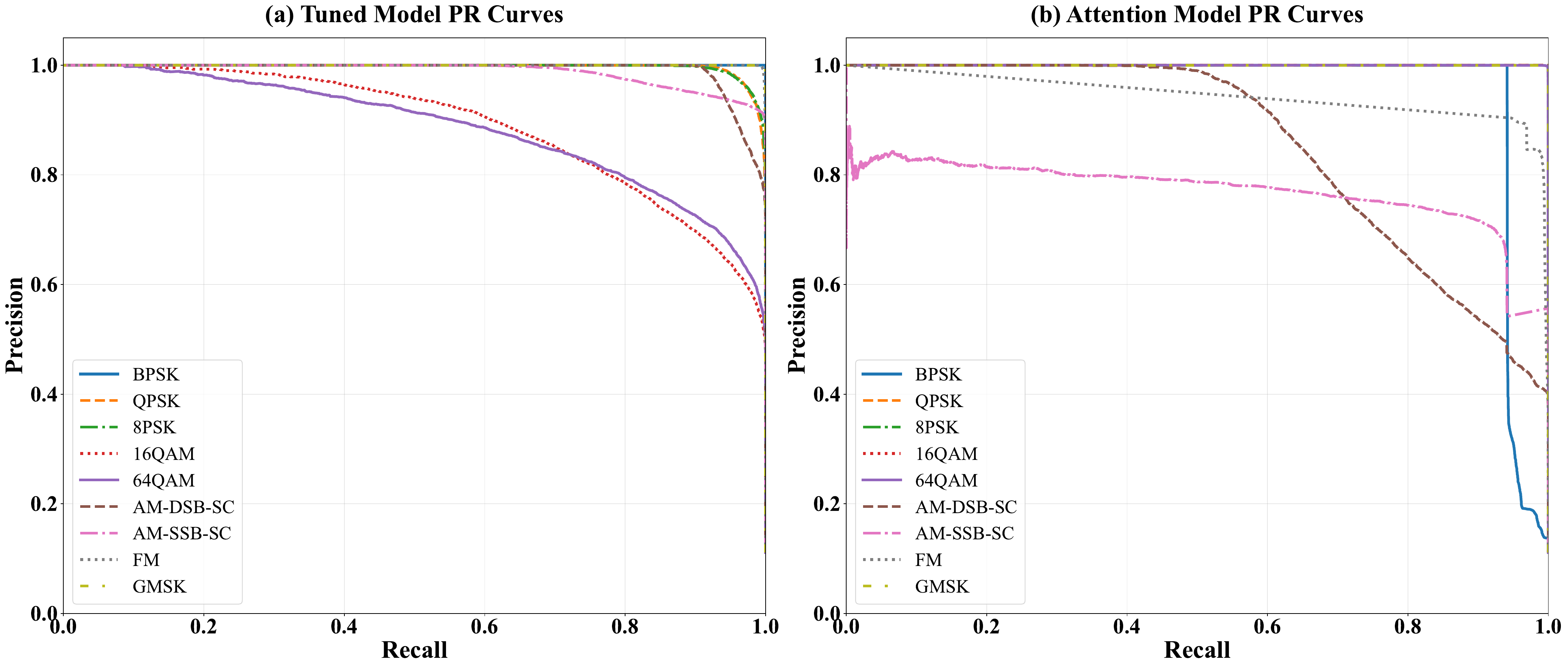}
    \caption{Precision-Recall curves illustrating the stability of the Proposed CNN-LSTM model (a) versus the degradation observed in the Attention model (b)}
    \label{fig:precision_recall_comparison}
\end{figure}

The PR curves depicted in Figure~\ref{fig:precision_recall_comparison} also mirrors these findings. For the CNN-LSTM Model (Figure~\ref{fig:precision_recall_comparison}(a)), the simplest modulation types hold great precision throughout the recall range. The complex classes 16QAM and 64QAM may experience a decrease in precision at high recall levels, however, the curves are still there and stable. The Attention Model (Figure~\ref{fig:precision_recall_comparison}(b)), on the other hand, experiences an extreme drop in precision for AM-SSB-SC and AM-DSB-SC that confirms the attention mechanism's difficulty in separating these signals from noise.

\begin{figure}[!t]
    \centering
    \includegraphics[width=\columnwidth]{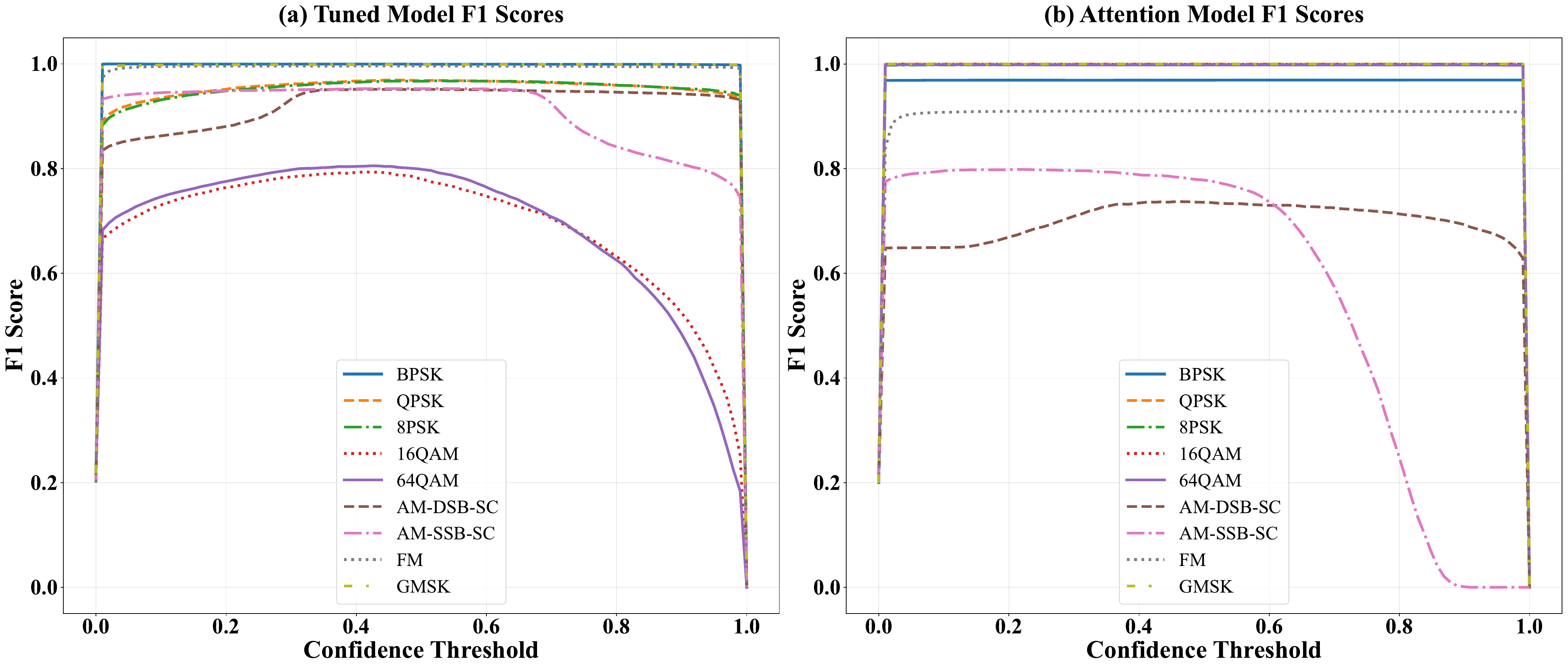}
    \caption{F1 score analysis highlighting the robustness of the Proposed CNN-LSTM model (a) compared to the instability of the Attention model (b)}
    \label{fig:f1_curve_comparison}
\end{figure}

Comparing the two models, the F1 score analysis in Figure~\ref{fig:f1_curve_comparison} reveals that the proposed model is more robust than the Attention-based one. The CNN-LSTM model that has been proposed keeps its accuracy at a high level and stable in all confidence thresholds (the curves with the flat top are the indication), whereas the Attention Model is quite unstable. The example of the same is the AM-SSB-SC class, performance of which drops drastically on increasing confidence thresholds, thus confirming that the proposed architecture is more adaptable to varying modulation schemes and noise conditions.

\begin{figure}[!t]
    \centering
    \includegraphics[width=\columnwidth]{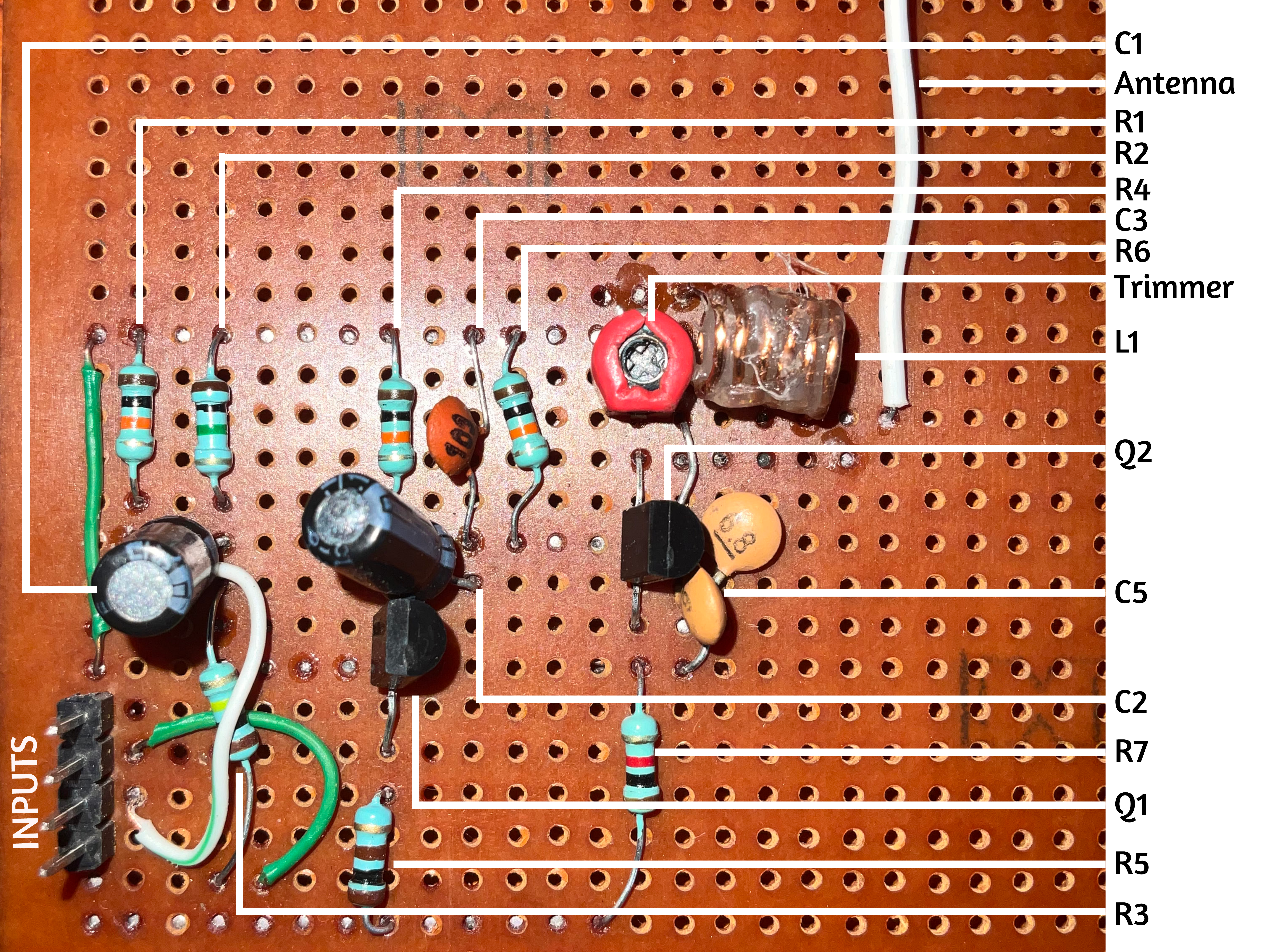}
    \caption{Frequency Modulation (FM) transmitter circuit schematic used for testing}
    \label{fig:fm_transmitter_circuit}
\end{figure}

To further ascertain real-world feasibility, we performed rigorous over-the-air (OTA) testing using a custom-built FM transmitter (Figure~\ref{fig:fm_transmitter_circuit}) and commercial broadcast signals. While most benchmark studies rely heavily on simulation, our approach integrates hardware implementation to confirm performance in live RF environments. We address this by training and testing on a uniform SNR distribution from 0 to 20~dB, covering both low and high noise scenarios. Despite this challenging range, our model achieves 93.48\% test accuracy, demonstrating robust performance across all SNR levels. This reflects the effectiveness of our training strategy, balanced dataset, and optimized CNN-LSTM configuration, which enable high performance even with heavily distorted signals. Compared to models reviewed in the literature, our method shows clear improvement. By ensuring general performance across the entire SNR range, rather than only at moderate or high SNRs, the proposed system is more reliable and better suited for real-world environments with variable noise levels.

\subsection{Limitations and Future Works}
While the results are promising, the limitations of the current approach must be acknowledged. The persistent confusion between high-order QAM schemes (16QAM and 64QAM) implies that the existing feature extraction may not fully capture the subtle differences in dense constellation patterns under noisy conditions. Furthermore, the model's performance degrades when transitioning from synthetic data to physical RF environments. On recorded real-world I/Q data, accuracy remained high at approximately 86\%. However, in a live, real-time environment, accuracy dropped to approximately 80\% due to operational complexities. This decay indicates that real-world factors, such as multipath fading, Doppler effects, and carrier frequency offsets, significantly impact classification performance.

The insufficient performance of the simplified attention mechanism, which compresses the 8-step sequence into a single vector, likely discards critical discriminative information. In contrast, our proposed model's method of flattening all hidden states preserves the complete, uncompressed temporal feature map. This allows the subsequent fully connected layers to acquire the complex patterns necessary for differentiating spectrally similar modulation schemes. This experiment validates our architectural decision, demonstrating that using the full temporal sequence is more beneficial for this task than a summarized attention-based context vector.

\section{Conclusion}
This research demonstrates the effectiveness of a hybrid CNN-LSTM architecture for AMC in realistic RF environments. The proposed system integrates SDR hardware with deep learning to achieve 93.48\% classification accuracy across nine modulation schemes, under challenging SNR conditions from 0 to 20~dB.

Our experimental findings highlight critical aspects of feature representation. In summary, the CNN component excels at capturing spatial patterns from I/Q data, while the LSTM layers model temporal dependencies. Most importantly, our comparative analysis demonstrated that preserving the full sequence of LSTM hidden states (via flattening) outperforms a temporal attention mechanism for this application. Although computationally faster, the attention-based method created an information bottleneck that severely degraded classification performance for complex analog modulations (AM-DSB-SC/SSB-SC). The ROC analysis confirmed near-perfect discrimination for most classes, with AUC values approaching 1.00.

Our approach achieves 93.48\% accuracy across this stringent 0--20~dB range, surpassing the approximately 90\% accuracy reported in similar studies. Notably, while many models report greater than 95\% accuracy at high SNR (greater than 14~dB), their performance degrades significantly below 10~dB. Our method maintains consistent performance across the entire range, indicating superior generalization.

The practical implications extend to cognitive radio, dynamic spectrum access, and military communications, where reliable modulation identification is critical. The demonstrated robustness across varying noise conditions, combined with SDR integration, positions this approach as a viable solution for deployment in real-world RF environments where signal quality is not guaranteed. The system's ability to maintain high accuracy under challenging conditions makes it particularly valuable for signal intelligence in contested electromagnetic environments.

\bibliographystyle{IEEEtran}
\bibliography{ref_rep}

\end{document}